\documentclass[11pt]{article}

\usepackage[final]{acl}

\usepackage{times}
\usepackage{latexsym}
\usepackage{booktabs}
\usepackage{amsmath} 

\usepackage{booktabs}
\usepackage[table]{xcolor}

\definecolor{headerblue}{RGB}{220,235,247}
\definecolor{tradblue}{RGB}{235,245,252}
\definecolor{llmorange}{RGB}{255,243,224}
\definecolor{bestgreen}{RGB}{220,242,220}

\usepackage[T1]{fontenc}

\usepackage[utf8]{inputenc}

\usepackage{microtype}

\usepackage{inconsolata}

\usepackage{graphicx}

\title{When Lexical Change Misleads: Rethinking Dynamic Topic Model Evaluation with Traditional and LLM-Based Metrics}

\author{Charu Karakkaparambil James \\
  RPTU University Kaiserslautern-Landau \\
Kaiserslautern, Germany\\
  \texttt{surname@cs.uni-kl.de} \\}

\begin{document}
\maketitle
\begin{abstract}
Dynamic topic models capture evolving word distributions, but traditional coherence metrics may fail when vocabulary changes while semantic meaning persists. We evaluate 120 topics from CoNTM and DLDA across NYT, DBLP, and arXiv, using three human annotators and Low, Medium, and High lexical-change categories. Traditional temporal coherence shows highly variable agreement with human judgments ($\rho$=-0.256 to 0.614). In contrast, LLM-based semantic similarity agrees strongly with human semantic judgments for CoNTM on NYT ($\rho$=0.609), DBLP ($\rho$=0.721), and arXiv ($\rho$=0.502), but is less consistent for DLDA. Lexical-change stratification reveals variation hidden by aggregate evaluation. We therefore advocate lexical-change-aware evaluation, jointly reporting traditional coherence and LLM-based semantic measures as complementary rather than interchangeable signals.
\end{abstract}

\section{Introduction}

Dynamic topic models (DTMs) extend conventional topic modeling by allowing latent topics to evolve across time \cite{sasaki2014online, iwata2009topic, yao2020tracking, karakkaparambil-james-etal-2026-semantic}. Instead of representing a topic using one static set of words, a dynamic topic may contain substantially different vocabulary at different time points while still describing a continuous underlying phenomenon. The original Dynamic Topic Model explicitly motivated this setting by noting that a scientific subject can look lexically different across decades even when the underlying intellectual trajectory remains connected \cite{10.1145/1143844.1143859}.

Evaluating such trajectories remains difficult. Topic-model evaluation has traditionally relied on automatic coherence measures based on statistical relationships between topic words. Human-centered work has repeatedly shown, however, that likelihood or automated coherence does not always correspond to human interpretability \cite{2984093.2984126,1857999.1858011}. More recent analyses have questioned whether coherence measures developed for classical topic models remain valid for newer models and across different application settings \cite{doogan-buntine-2021-topic,3540261.3540416}.

The problem is amplified in temporal topic modeling. Consider a topic whose early vocabulary concerns desktop, computer, software, and internet, while later years emphasize smartphone, app, cloud, and platform. Lexically, the topic has changed considerably. Semantically, however, a human reader may recognize a coherent evolution of digital technology. A metric dominated by lexical co-occurrence can therefore answer a different question from the one a human asks when interpreting a temporal narrative.

Large language models provide a possible complementary signal because they can reason over the semantic relationship between different surface realizations of a topic. \citet{stammbach-etal-2023-revisiting}, for example, showed that LLM judgments can correlate strongly with human assessments in static topic-model evaluation, although the effectiveness depends on the evaluation task. We extend this motivation to temporal topic evolution.

We study two models, CoNTM \cite{karakkaparambil-james-etal-2026-continual} and DLDA \cite{10.1145/1143844.1143859}, on three temporally structured corpora. Our main finding is that metric validity is conditional. Traditional coherence correlates strongly with human coherence in some conditions but weakly or even negatively in others. Similarly, LLM semantic evaluation performs strongly for CoNTM but is inconsistent for DLDA. Lexical-change stratification reveals patterns that aggregate evaluation obscures.

\begin{figure*}[htp!]
    \centering
    \begin{minipage}[b]{1.00\textwidth}
        \centering
        \includegraphics[width=\textwidth]{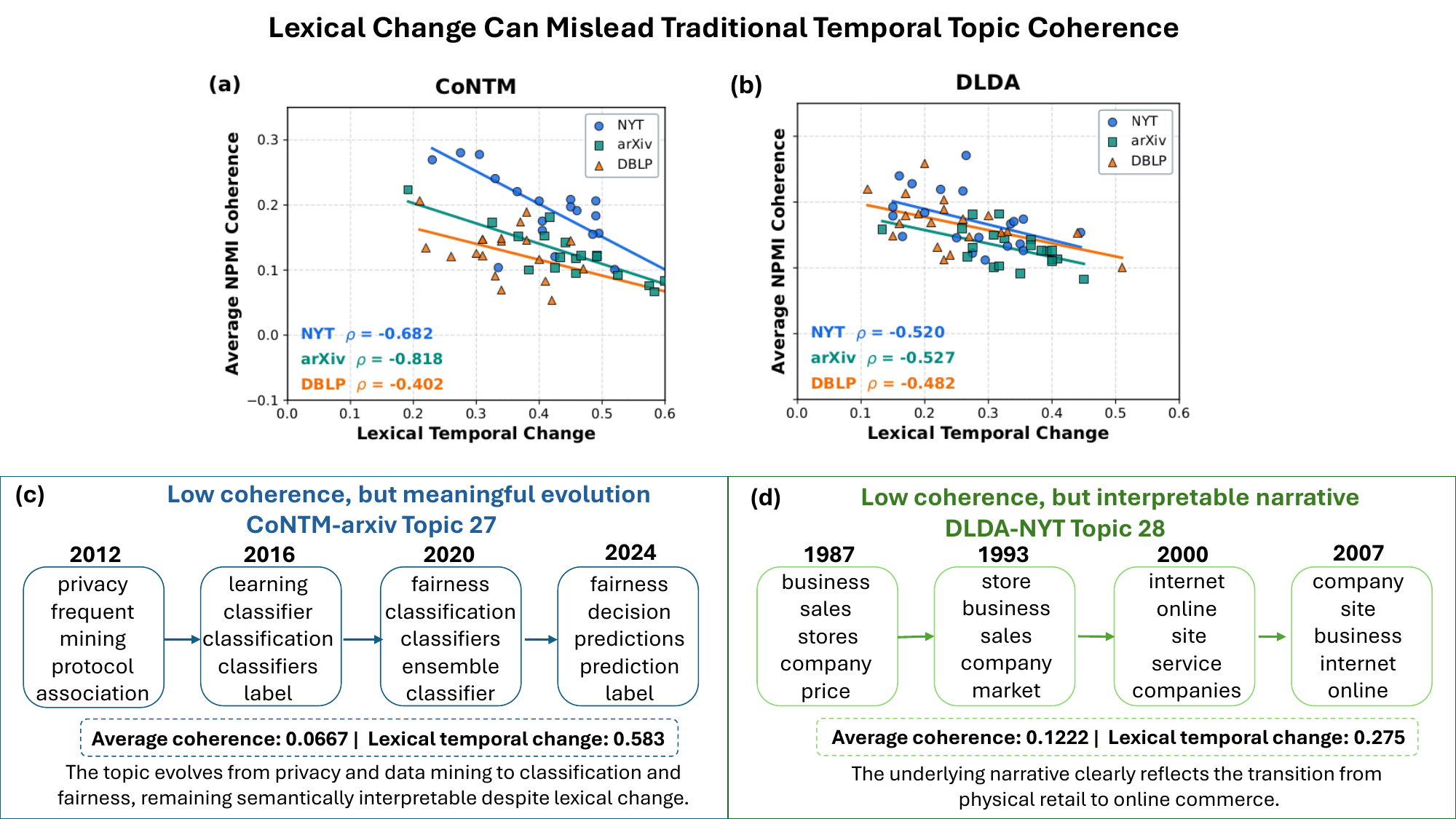}
        \caption{Relationship between lexical temporal change and traditional NPMI coherence for CoNTM and DLDA across NYT, arXiv, and DBLP. Panels (a–b) show a negative association between lexical change and coherence across datasets. Panels (c–d) illustrate that topics with relatively low coherence can nevertheless exhibit meaningful and interpretable semantic evolution, motivating evaluation beyond lexical coherence alone.}
        \label{fig:lexicalChange}
        \vspace{-0.6cm}
    \end{minipage}
\vspace{-0.3cm}
\end{figure*}
Our main contributions are:
\begin{itemize}
    \item We analyze \textbf{traditional temporal coherence} across two dynamic topic models and three datasets, showing substantial variation in agreement with human judgments.

    \item We introduce \textbf{lexical-change-aware evaluation} by grouping topics into Low, Medium, and High lexical-change regimes, revealing patterns hidden by aggregate correlations.

    \item We evaluate \textbf{LLM-based semantic similarity} against human judgments, showing that it provides a complementary semantic signal but is not uniformly reliable.

    \item We advocate a \textbf{multi-view evaluation framework} combining traditional coherence, LLM-based semantic similarity, and lexical change.
\end{itemize}

This paper is organized as follows. Section \ref{Related Work} reviews related work, Section \ref{Experimental Setup} describes the experimental setup, Section \ref{Results} presents the results, Section \ref{Discussion} discusses the findings, and Section \ref{Conclusion} concludes the paper.

\section{Related Work}
\label{Related Work}

This section reviews prior work on traditional topic-model evaluation and recent LLM-based approaches to semantic evaluation, with a focus on their relevance to dynamic topic models.

\paragraph{Topic model evaluation} Traditional topic-model evaluation commonly relies on coherence measures based on statistical associations among topic words. However, automated metrics do not always align with human interpretability \cite{2984093.2984126,1857999.1858011}. More recent studies show that this mismatch can persist across neural topic models and different domains, questioning the use of coherence as a universal proxy for human judgment \cite{doogan-buntine-2021-topic,3540261.3540416}. This issue is especially relevant for dynamic topic models, where vocabulary can change substantially over time while the underlying semantic theme remains coherent.

\paragraph{LLM-based evaluation} Recent work has explored LLMs as semantic evaluators of topic-model outputs.  \citet{stammbach-etal-2023-revisiting} show that LLM judgments can correlate strongly with human topic evaluations, although performance depends on the evaluation formulation. Existing work primarily considers static topic representations \cite{karakkaparambil-james-etal-2024-evaluating}. In contrast, we examine temporal topic evolution and ask whether the reliability of traditional and LLM-based metrics changes with Low, Medium, and High lexical change. Rather than replacing traditional coherence, we study traditional coherence and LLM-based semantic similarity as complementary signals for evaluating dynamic topics.

\section{Experimental Setup}
\label{Experimental Setup}

This section describes the models and datasets, human evaluation procedure, traditional and LLM-based metrics, and lexical-change analysis used in our experiments.

\subsection{Models and datasets}
We evaluate CoNTM and DLDA on three temporally structured datasets: the New York Times (NYT), DBLP, and arXiv. For each model–dataset combination, we select 20 topics, resulting in 120 temporal topic trajectories in total. Each topic is represented at three time points: 1987, 1997, and 2007 for NYT \cite{AB2/GZC6PL_2008}; 2010, 2015, and 2020 for DBLP \cite{ley2002dblp}; and 2012, 2018, and 2024 for arXiv \cite{arxiv_org_submitters_2023}.

\subsection{Human evaluation}
Three annotators evaluate every topic. For consistency, we use three responses for all model–dataset combinations; the DLDA–NYT file contains five responses, from which the first three are used in the present experiment.

Annotators answer five questions on a five-point scale:

\begin{itemize}
    \item Temporal Topic Coherence: Are the topic words coherent and interpretable when consecutive time points are considered?
    \item Temporal Topic Smoothness: Does the topic evolve smoothly between the three temporal points?
    \item Beginning Theme Accuracy: Does the generated beginning theme accurately represent the initial topic words?
    \item Ending Theme Accuracy: Does the generated ending theme accurately represent the final topic words?
    \item Semantic Similarity: How semantically similar are the beginning and ending themes?
\end{itemize}

This yields (120$\times$3$\times$5=1{,}800) scalar human judgments. We average the three annotators for each topic.

The present analysis focuses primarily on Temporal Topic Coherence and Semantic Similarity because these correspond directly to the traditional and LLM-based metrics studied below.

\subsection{Traditional and LLM-Based Evaluation}

For traditional evaluation, we use the automatically computed \textit{Temporal Topic Coherence} \cite{karakkaparambil-james-etal-2024-evaluating}, which summarizes topic coherence across the temporal trajectory. We compare this score with the mean human \textit{Temporal Topic Coherence} (TTC) rating obtained from the three annotators.


For semantic evaluation, we use GPT-5.5 \cite{openai2026gpt55} through the OpenAI Responses API. For each temporal topic pair, the model receives the dataset, dynamic topic model, topic identifier, timestamps, and corresponding top topic words. Using a fixed prompt, GPT-5.5 generates concise semantic descriptions of the two topic states and evaluates their semantic similarity on a five-point scale, ranging from 1 (semantically unrelated) to 5 (semantically equivalent). The prompt explicitly instructs the model to distinguish genuine semantic change from lexical replacement and not to penalize vocabulary change when the underlying topic remains conceptually similar. The model returns a structured JSON response containing the two semantic themes, a similarity score, and a justification. We compare this score with the mean human \textit{Semantic Similarity} (SS) rating.

Importantly, the traditional and LLM-based metrics are evaluated against different human judgments. Specifically, we examine

\begin{equation}
C_{\mathrm{trad.}}
\leftrightarrow
H_{\mathrm{TTC}},
\end{equation}

and

\begin{equation}
S_{\mathrm{LLM}}
\leftrightarrow
H_{\mathrm{SS}},
\end{equation}

where $C_{\mathrm{trad.}}$ denotes temporal topic coherence, $S_{\mathrm{LLM}}$ denotes LLM-based semantic similarity, $H_{\mathrm{TTC}}$ denotes the mean human temporal topic coherence judgment, and $H_{\mathrm{SS}}$ denotes the mean human semantic similarity judgment. Therefore, we do not interpret the two correlation coefficients as a direct comparison between competing metrics. Instead, we assess how well each metric agrees with human judgments of the
corresponding property of temporal topic quality.

\subsection{Lexical-change stratification}

Each topic has a Topic Lexical Temporal Change value and an associated Low, Medium, or High lexical-change category. Within each model–dataset combination, the 20 topics consist of seven Low, seven Medium, and six High lexical-change topics.

This stratification allows us to test whether aggregate metric–human correlations conceal different behavior under different levels of vocabulary evolution.

Because human responses are ordinal Likert judgments, Spearman's rank correlation ($\rho$) is used as the primary agreement statistic. Category-level results should be interpreted cautiously because each category contains only six or seven topics per dataset.

\section{Results}
\label{Results}

This section presents the overall and lexical change analyzed correlations between automatic metrics and their corresponding human judgments, highlighting how metric reliability varies across models and datasets.

\subsection{Traditional coherence does not have uniform human validity}
Table \ref{tab:overall_correlation} reports the relationship between traditional temporal coherence and human Temporal Topic Coherence, together with the relationship between the LLM score and human Semantic Similarity.

\begin{table}[ht!]
\centering
\small
\setlength{\tabcolsep}{2.5pt}
\renewcommand{\arraystretch}{1.12}

\begin{tabular}{llcc}
\toprule

\rowcolor{headerblue}
\textbf{Model} &
\textbf{Dataset} &
\textbf{$C_{\mathrm{trad.}} \leftrightarrow H_{\mathrm{TTC}}$} &
\textbf{$S_{\mathrm{LLM}}\leftrightarrow H_{\mathrm{SS}}$} \\

\midrule

CoNTM & NYT &
\cellcolor{tradblue}$-0.256$ &
\cellcolor{bestgreen}\textbf{$0.609$} \\

CoNTM & DBLP &
\cellcolor{tradblue}$0.425$ &
\cellcolor{bestgreen}\textbf{$0.721$} \\

CoNTM & arXiv &
\cellcolor{bestgreen}\textbf{$0.500$} &
\cellcolor{bestgreen}\textbf{$0.502$} \\

\midrule

DLDA & NYT &
\cellcolor{bestgreen}\textbf{$0.531$} &
\cellcolor{bestgreen}\textbf{$0.445$} \\

DLDA & DBLP &
\cellcolor{bestgreen}\textbf{$0.614$} &
\cellcolor{llmorange}$-0.086$ \\

DLDA & arXiv &
\cellcolor{tradblue}$0.007$ &
\cellcolor{llmorange}$-0.190$ \\

\bottomrule
\end{tabular}

\caption{Spearman's $\rho$ between automatic metrics and corresponding three-annotator human judgments. $C_{\mathrm{trad.}}$ denotes traditional temporal coherence, $H_{\mathrm{TTC}}$ denotes human Temporal Topic Coherence, and $H_{\mathrm{SS}}$ denotes human Semantic Similarity. Bold values indicate statistically significant correlations
($p<0.05$).}

\label{tab:overall_correlation}
\vspace{-0.3cm}
\end{table}

Traditional temporal coherence behaves very differently across experimental conditions. For DLDA–DBLP, it agrees strongly with humans ($\rho=0.614,p=.004$); DLDA–NYT also exhibits significant positive agreement ($\rho=0.531,p=.016$). CoNTM–arXiv produces moderate agreement ($\rho=0.500,p=.025$).

However, these results do not generalize across datasets. For DLDA–arXiv, correlation is essentially zero ($\rho=0.007$). More strikingly, CoNTM–NYT shows a negative overall correlation ($\rho=-0.256$), although it is not statistically significant.

Thus, a high traditional coherence score cannot be assumed to have the same relationship with human temporal interpretability across models and corpora. This supports prior concerns that automated topic coherence is not universally interchangeable with human judgment \cite{doogan-buntine-2021-topic,3540261.3540416}.

\subsection{Lexical change reveals hidden metric behavior}
The aggregate results hide substantial differences among lexical-change categories. Particularly notable examples include:

\begin{table*}[t]
\centering
\small
\setlength{\tabcolsep}{10pt}
\renewcommand{\arraystretch}{1.15}

\begin{tabular}{lccc}
\toprule

\rowcolor{headerblue}
\textbf{Setting} &
\textbf{Low} &
\textbf{Medium} &
\textbf{High} \\

\midrule

\rowcolor{tradblue}
CoNTM--NYT, 
$C_{\mathrm{trad.}} \leftrightarrow H_{\mathrm{TTC}}$
&
\cellcolor{bestgreen}\textbf{$-0.786$}
& $0.116$
& $0.516$ \\

\rowcolor{tradblue}
DLDA--NYT,
$C_{\mathrm{trad.}} \leftrightarrow H_{\mathrm{TTC}}$
&
$0.094$
& \cellcolor{bestgreen}\textbf{$0.873$}
& $0.278$ \\

\rowcolor{tradblue}
DLDA--DBLP,
$C_{\mathrm{trad.}} \leftrightarrow H_{\mathrm{TTC}}$
&
$0.177$
& \cellcolor{bestgreen}\textbf{$0.898$}
& $0.359$ \\

\midrule

\rowcolor{llmorange}
CoNTM--NYT,
$S_{\mathrm{LLM}}$ $\leftrightarrow H_{\mathrm{SS}}$
&
\cellcolor{bestgreen}\textbf{$0.801$}
& $0.750$
& $-0.016$ \\

\rowcolor{llmorange}
CoNTM--DBLP,
$S_{\mathrm{LLM}}$ $\leftrightarrow H_{\mathrm{SS}}$
&
$0.600$
& \cellcolor{bestgreen}\textbf{$0.786$}
& $0.626$ \\

\rowcolor{llmorange}
CoNTM--arXiv,
$S_{\mathrm{LLM}}$ $\leftrightarrow H_{\mathrm{SS}}$
&
$0.730$
& $0.311$
& $0.117$ \\

\bottomrule
\end{tabular}

\caption{
Selected lexical-change-stratified Spearman correlations ($\rho$). Blue rows represent traditional coherence versus human Temporal Topic
Coherence (TTC), while orange rows represent LLM semantic similarity
versus human Semantic Similarity (SS). Green cells highlight the
statistically significant correlations ($p<0.05$).
Category sizes are $N=7$, $7$, and $6$ for Low, Medium, and High
lexical change, respectively.
}
\label{tab:lexical_stratified}
\vspace{-0.5cm}
\end{table*}

The strongest example of metric instability occurs for CoNTM–NYT. Under Low lexical change, traditional coherence is strongly negatively correlated with human coherence ($\rho=-0.786,p=.036$), whereas the LLM semantic similarity score strongly agrees with human semantic judgments in the same lexical regime ($\rho=0.801,p=.030$).

DLDA shows a different pattern. Traditional coherence is particularly informative for Medium lexical change, reaching $\rho=0.873$ on NYT and $\rho=0.898$ on DBLP. This observation is important because it prevents an overly simple conclusion that increasing lexical change necessarily invalidates traditional coherence. Instead, the effect depends on the interaction among lexical change, model behavior, and corpus characteristics.

\subsection{LLM semantic evaluation is useful but not universal}

Across all 120 topics, human Semantic Similarity and the LLM semantic score have a pooled correlation of
$$[
\rho = 0.479,\qquad p < .001.
]$$

The pooled correlation is strongest for Low lexical-change topics ($\rho = 0.556, p < .001$), remains significant for Medium lexical change ($\rho = 0.405, p = .008$), and becomes weaker for High lexical change ($\rho = 0.294, p = .082$).

Model-specific analysis reveals an important asymmetry. Across the three datasets, CoNTM has strong LLM–human semantic agreement:
$$[
\rho_{\text{NYT}} = 0.609,
\rho_{\text{DBLP}} = 0.721, \\
\rho_{\text{arXiv}} = 0.502.
]$$

All three are statistically significant.

DLDA is substantially less consistent. NYT shows moderate agreement ($\rho = 0.445, p = .049$), but DBLP ($\rho = -0.086$) and arXiv ($\rho = -0.190$) do not.

These results argue against replacing traditional metrics with an LLM score. Instead, they suggest that LLM judgments form a second evaluation view whose reliability must itself be validated. This is consistent with existing LLM-based topic-model evaluation work showing that performance depends strongly on the evaluation formulation.

\section{Discussion: When Does Lexical Change Mislead?}
\label{Discussion}

Our results (see Table \ref{tab:lexical_stratified}) suggest that lexical change should be treated as an evaluation context variable, rather than merely as another topic-quality score.

Traditional coherence asks whether topic words exhibit statistical or lexical compatibility. This is useful when lexical association is closely aligned with the semantic organization perceived by humans. DLDA–NYT and DLDA–DBLP demonstrate that this alignment can be very strong.

Semantic continuity asks a different question: After the words change, is the topic still about the same underlying theme?

An LLM can compare descriptions such as a Beginning Theme and Ending Theme without requiring the same words to persist. This capability is particularly valuable when semantic relationships are not reducible to direct lexical overlap.

However, the results also show that lexical evolution is not equivalent to semantic evolution. High lexical change does not automatically imply that a traditional metric fails, nor does it guarantee that an LLM succeeds. Indeed, the LLM–human relationship weakens for several High-change conditions. This may occur because highly changed trajectories are genuinely ambiguous: both humans and LLMs must determine whether the trajectory represents meaningful evolution, topic drift, or a transition between unrelated concepts.

We therefore propose evaluating a dynamic topic using three complementary dimensions:

$$
[
\boxed{
E(T)=
{
C_{\text{traditional}},
S_{\text{LLM}},
L_{\text{change}}
}
}
]$$

where ($C_{\text{traditional}}$) measures lexical/statistical coherence, ($S_{\text{LLM}}$) measures semantic continuity, and ($L_{\text{change}}$) describes the amount of vocabulary evolution.

Crucially, we do not collapse these values into a single weighted score in the present work. The human experiments show that their behavior changes across models and lexical regimes; choosing fixed weights without further validation could simply introduce another opaque evaluation metric.

Instead, we recommend reporting them jointly. For example, a topic with low traditional coherence but high semantic similarity and high lexical change should be inspected differently from a topic that scores poorly on both coherence and semantic similarity. The first may represent legitimate semantic evolution; the second is more likely to represent an unstable or incoherent topic trajectory.

This interpretation leads to the paper's main conclusion that Lexical change does not make traditional evaluation useless; it makes reliance on a single evaluation unsafe.

\section{Conclusion}
\label{Conclusion}

We investigated the relationship among lexical change, traditional coherence, LLM-based semantic similarity, and human judgments in dynamic topic-model evaluation. Across CoNTM and DLDA on NYT, DBLP, and arXiv, traditional coherence exhibits substantial variation in human agreement, including strong positive, near-zero, and negative correlations. LLM-based semantic similarity provides strong complementary evidence for several CoNTM conditions but is itself model dependent.

These findings suggest that dynamic topic evaluation should move beyond the search for one universal automatic metric. A temporal topic is simultaneously a lexical object and a semantic trajectory. Traditional coherence captures whether its words fit together; LLM-based evaluation captures whether its meaning persists or evolves coherently. Lexical change tells us when the distinction matters.

We therefore advocate lexical-change-aware, multi-view evaluation in which traditional and LLM-based metrics are jointly reported and validated against their corresponding human judgments.

\section*{Limitations}
Our study evaluates 120 topics from two models and three English-language datasets. Category-level experiments contain only six or seven topics, making individual correlations sensitive to outliers. The LLM analysis uses one LLM configuration, and LLM judgments may vary with prompting, model family, and generated semantic themes. In addition, traditional coherence and LLM semantic similarity are validated against different human questions, so their correlation coefficients should not be interpreted as direct metric rankings.

Future work should increase the number of topics and annotators, evaluate additional temporal topic models and LLMs, test prompt robustness, and investigate whether lexical-change-aware combinations predict holistic human judgments better than any individual metric.



\bibliography{custom}

@inproceedings{10.1145/1143844.1143859,
author = {Blei, David M. and Lafferty, John D.},
title = {Dynamic topic models},
year = {2006},
isbn = {1595933832},
publisher = {Association for Computing Machinery},
address = {New York, NY, USA},
url = {https://doi.org/10.1145/1143844.1143859},
doi = {10.1145/1143844.1143859},
booktitle = {Proceedings of the 23rd International Conference on Machine Learning},
pages = {113–120},
numpages = {8},
location = {Pittsburgh, Pennsylvania, USA},
series = {ICML '06}
}

@inproceedings{2984093.2984126,
author = {Chang, Jonathan and Boyd-Graber, Jordan and Gerrish, Sean and Wang, Chong and Blei, David M.},
title = {Reading tea leaves: how humans interpret topic models},
year = {2009},
isbn = {9781615679119},
publisher = {Curran Associates Inc.},
address = {Red Hook, NY, USA},
booktitle = {Proceedings of the 23rd International Conference on Neural Information Processing Systems},
pages = {288–296},
numpages = {9},
location = {Vancouver, British Columbia, Canada},
series = {NIPS'09}
}

@inproceedings{1857999.1858011,
author = {Newman, David and Lau, Jey Han and Grieser, Karl and Baldwin, Timothy},
title = {Automatic evaluation of topic coherence},
year = {2010},
isbn = {1932432655},
publisher = {Association for Computational Linguistics},
address = {USA},
booktitle = {Human Language Technologies: The 2010 Annual Conference of the North American Chapter of the Association for Computational Linguistics},
pages = {100–108},
numpages = {9},
location = {Los Angeles, California},
series = {HLT '10}
}

@inproceedings{doogan-buntine-2021-topic,
    title = "Topic Model or Topic Twaddle? Re-evaluating Semantic Interpretability Measures",
    author = "Doogan, Caitlin  and
      Buntine, Wray",
    editor = "Toutanova, Kristina  and
      Rumshisky, Anna  and
      Zettlemoyer, Luke  and
      Hakkani-Tur, Dilek  and
      Beltagy, Iz  and
      Bethard, Steven  and
      Cotterell, Ryan  and
      Chakraborty, Tanmoy  and
      Zhou, Yichao",
    booktitle = "Proceedings of the 2021 Conference of the North American Chapter of the Association for Computational Linguistics: Human Language Technologies",
    month = jun,
    year = "2021",
    address = "Online",
    publisher = "Association for Computational Linguistics",
    url = "https://aclanthology.org/2021.naacl-main.300/",
    doi = "10.18653/v1/2021.naacl-main.300",
    pages = "3824--3848"
}

@inproceedings{3540261.3540416,
author = {Hoyle, Alexander and Goel, Pranav and Peskov, Denis and Hian-Cheong, Andrew and Boyd-Graber, Jordan and Resnik, Philip},
title = {Is automated topic model evaluation broken? the incoherence of coherence},
year = {2021},
isbn = {9781713845393},
publisher = {Curran Associates Inc.},
address = {Red Hook, NY, USA},
booktitle = {Proceedings of the 35th International Conference on Neural Information Processing Systems},
articleno = {155},
numpages = {16},
series = {NIPS '21}
}

@inproceedings{stammbach-etal-2023-revisiting,
    title = "Revisiting Automated Topic Model Evaluation with Large Language Models",
    author = "Stammbach, Dominik  and
      Zouhar, Vil{\'e}m  and
      Hoyle, Alexander  and
      Sachan, Mrinmaya  and
      Ash, Elliott",
    editor = "Bouamor, Houda  and
      Pino, Juan  and
      Bali, Kalika",
    booktitle = "Proceedings of the 2023 Conference on Empirical Methods in Natural Language Processing",
    month = dec,
    year = "2023",
    address = "Singapore",
    publisher = "Association for Computational Linguistics",
    url = "https://aclanthology.org/2023.emnlp-main.581/",
    doi = "10.18653/v1/2023.emnlp-main.581",
    pages = "9348--9357"
}

@inproceedings{karakkaparambil-james-etal-2026-continual,
    title = "Continual Neural Topic Model",
    author = "Karakkaparambil James, Charu  and
      Mustafa, Waleed  and
      Monteiro, Marcio  and
      Kloft, Marius  and
      Fellenz, Sophie",
    editor = "Demberg, Vera  and
      Inui, Kentaro  and
      Marquez, Llu{\'i}s",
    booktitle = "Proceedings of the 19th Conference of the {E}uropean Chapter of the {A}ssociation for {C}omputational {L}inguistics (Volume 1: Long Papers)",
    month = mar,
    year = "2026",
    address = "Rabat, Morocco",
    publisher = "Association for Computational Linguistics",
    url = "https://aclanthology.org/2026.eacl-long.312/",
    doi = "10.18653/v1/2026.eacl-long.312",
    pages = "6636--6658",
    ISBN = "979-8-89176-380-7"
}

@inproceedings{karakkaparambil-james-etal-2024-evaluating,
    title = "Evaluating Dynamic Topic Models",
    author = "Karakkaparambil James, Charu  and
      Nagda, Mayank  and
      Haji Ghassemi, Nooshin  and
      Kloft, Marius  and
      Fellenz, Sophie",
    editor = "Ku, Lun-Wei  and
      Martins, Andre  and
      Srikumar, Vivek",
    booktitle = "Proceedings of the 62nd Annual Meeting of the Association for Computational Linguistics (Volume 1: Long Papers)",
    month = aug,
    year = "2024",
    address = "Bangkok, Thailand",
    publisher = "Association for Computational Linguistics",
    url = "https://aclanthology.org/2024.acl-long.11/",
    doi = "10.18653/v1/2024.acl-long.11",
    pages = "160--176"
}

@inproceedings{sasaki2014online,
  title={Online topic model for twitter considering dynamics of user interests and topic trends},
  author={Sasaki, Kentaro and Yoshikawa, Tomohiro and Furuhashi, Takeshi},
  booktitle={Proceedings of the 2014 conference on empirical methods in natural language processing (EMNLP)},
  pages={1977--1985},
  year={2014}
}

@inproceedings{iwata2009topic,
  title={Topic tracking model for analyzing consumer purchase behavior},
  author={Iwata, Tomoharu and Watanabe, Shinji and Yamada, Takeshi and Ueda, Naonori},
  booktitle={Twenty-First international joint conference on artificial intelligence},
  year={2009}
}

@article{yao2020tracking,
  title={Tracking urban geo-topics based on dynamic topic model},
  author={Yao, Fang and Wang, Yan},
  journal={Computers, Environment and Urban Systems},
  volume={79},
  pages={101419},
  year={2020},
  publisher={Elsevier}
}

@inproceedings{karakkaparambil-james-etal-2026-semantic,
    title = "Semantic Echo Pathways ({SEP}): Tracing How Medical Language Propagates and Transforms",
    author = "Karakkaparambil James, Charu  and
      Monteiro, Marcio  and
      Fellenz, Sophie",
    editor = {Danilova, Vera  and
      Kurfal{\i}, Murathan  and
      S{\"o}derfeldt, Ylva  and
      Reed, Julia  and
      Burchell, Andrew},
    booktitle = "Proceedings of the 1st Workshop on Linguistic Analysis for Health ({H}ea{L}ing 2026)",
    month = mar,
    year = "2026",
    address = "Rabat, Morocco",
    publisher = "Association for Computational Linguistics",
    url = "https://aclanthology.org/2026.healing-1.5/",
    doi = "10.18653/v1/2026.healing-1.5",
    pages = "55--66",
    ISBN = "979-8-89176-367-8"
}

@misc{AB2/GZC6PL_2008,
author = {Sandhaus, Evan},
publisher = {Abacus Data Network},
title = {{The New York Times Annotated Corpus}},
year = {2008},
version = {V1},
doi = {11272.1/AB2/GZC6PL},
url = {https://hdl.handle.net/11272.1/AB2/GZC6PL}
}

@misc{arxiv_org_submitters_2023,
	title={arXiv Dataset},
	url={https://www.kaggle.com/dsv/4852963},
	DOI={10.34740/KAGGLE/DSV/4852963},
	publisher={Kaggle},
	author={arXiv.org submitters},
	year={2023}
}

@inproceedings{ley2002dblp,
  title={The DBLP computer science bibliography: Evolution, research issues, perspectives},
  author={Ley, Michael},
  booktitle={International symposium on string processing and information retrieval},
  pages={1--10},
  year={2002},
  organization={Springer}
}

@misc{openai2026gpt55,
  author = {{OpenAI}},
  title = {Introducing {GPT-5.5}},
  year = {2026},
  howpublished = {OpenAI},
}




\end{document}